\documentclass[lang=english]{tumarxivarticle}

\usepackage[
    backend=biber,
    style=numeric,
    maxbibnames=99,
    maxcitenames=1,
    doi=false,
    isbn=false,
    url=false,
    date=year,
    sorting=nyt
  ]{biblatex}
\newcommand{\citep}{\parencite}

\newcommand{\citet}{\textcite}
\newcommand{\Citet}{\Textcite}

\bibliography{bib.bib}

\newcommand{\loss}{\ensuremath{\mathcal{L}}}

\usepackage[shortcuts]{extdash}
\usepackage{eso-pic}

\begin{document}


\AddToShipoutPicture*{%
  \AtTextUpperLeft{%
    \put(0,\LenToUnit{.75cm}){%
      \fbox{\parbox[][.4cm][c]{\textwidth}{%
        \centering
        This work was accepted for presentation at the IEEE Intelligent Transportation Systems Conference (ITSC) 2019.
      }}%
    }%
  }%
}%

\title{
  \normalfont
  MultiDepth: Single-Image Depth Estimation via Multi-Task Regression and Classification
  }

\setkomafont{author}{\large}
\author{
  Lukas Liebel, Marco Körner
  }
\publishers{\normalsize
  Computer Vision Research Group, Chair of Remote Sensing Technology \\ Technical University of Munich, Germany \\ lukas.liebel@tum.de \\[0.25cm]
}
\date{}

\twocolumn[
  \maketitle
  \renewcommand{\abstractname}{}
  \begin{onecolabstract}

    We introduce MultiDepth, a novel training strategy and \gls{cnn} architecture that allows approaching \gls{side} as a \gls{mt} problem.
    \Gls{side} is an important part of \gls{rsu}.
    It, thus, plays a vital role in \acrlongpl{adas} and \glspl{av}.
    Best results for the \gls{side} task so far have been achieved using deep \glspl{cnn}.
    However, optimization of regression problems, such as estimating depth, is still a challenging task.
    For the related tasks of image classification and semantic segmentation, numerous \gls{cnn}-based methods with robust training behavior have been proposed.
    Hence, in order to overcome the notorious instability and slow convergence of depth value regression during training, MultiDepth makes use of depth interval classification as an auxiliary task.
    The auxiliary task can be disabled at test-time to predict continuous depth values using the main regression branch more efficiently.
    We applied MultiDepth to road scenes and present results on the \gls{kside} dataset.
    In experiments, we were able to show that end-to-end \gls{mt} learning with both, regression and classification, is able to considerably improve training and yield more accurate results.

  \end{onecolabstract}
  \vspace{1cm}
]

\glsresetall
\defglsentryfmt{\ifglsused{\glslabel}{\glsgenentryfmt}{\emph{\glsgenentryfmt}}}

\flushbottom 


\section{Introduction}

\begin{figure}[t!]
  \includegraphics[trim=0 0 0 110,clip,width=\linewidth]{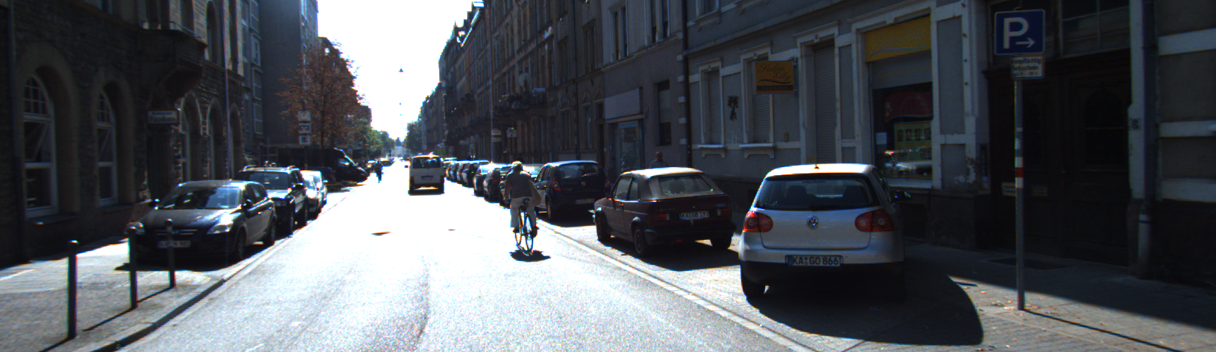} \\[2mm]
  \includegraphics[trim=0 0 0 110,clip,width=\linewidth]{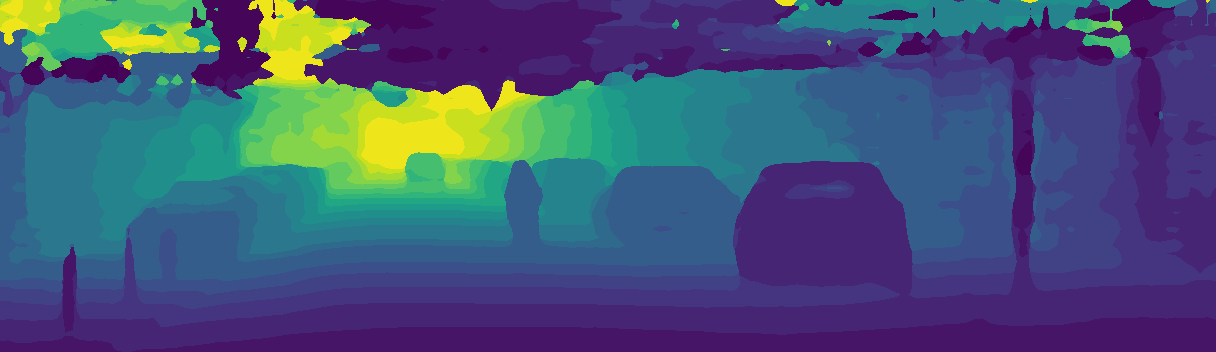} \\[2mm]
  \includegraphics[trim=0 0 0 110,clip,width=\linewidth]{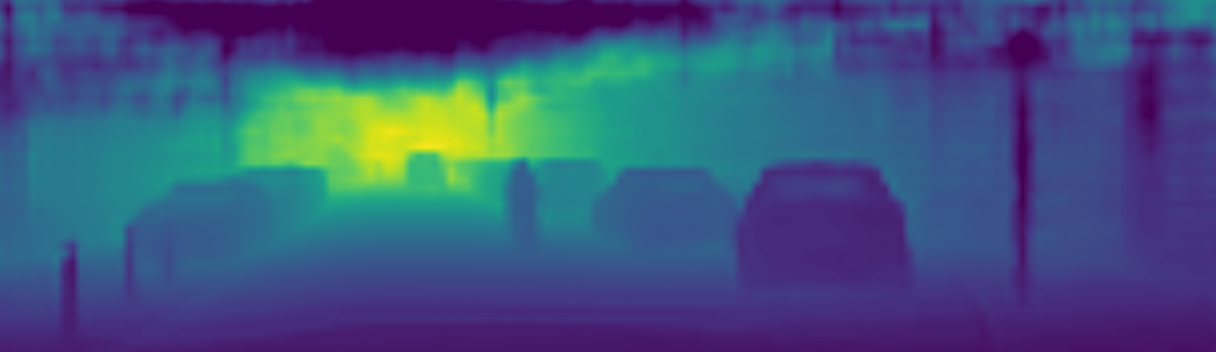}
  \caption{Depth prediction (bottom) for a single RGB image (top) from the KITTI depth prediction benchmark \citep{Uhrig17, Geiger13}.
  Auxiliary depth interval classification result (middle) used during training to support the depth value regression via optimization of a multi-task objective.}
  \label{fig:intro_example}
  \vspace{-0.4cm}
\end{figure}

Depth estimation is an important part of scene understanding in various domains.
Traditionally, depth maps are derived from active sensor measurements, such as \gls{lidar} point clouds, or from stereo images.
However, in the absence of observations allowing for the explicit reconstruction of pixel-wise depth values for a corresponding image, methods for directly estimating depth from a single monocular image are required.
A typical application lies in robotics, most prominently \glspl{av}, where a high degree of redundancy is of vital importance.
\Cref{fig:intro_example} exemplarily shows the result of \gls{side} for a road scene.

Predicting depth from a single image has seen substantial improvements due to the rise of \gls{dl}-based methods.
First approaches to \gls{side} for indoor scenes using deep \glspl{cnn} were presented by \citet{Eigen14,Eigen15}.
Ever since then, various methods for the prediction \citep{Laina16,Liu15,Li17,Liu18,Heo18,Zioulis18} and evaluation \citep{Koch18} of depth maps for indoor scenes have been proposed.

\Gls{side} in unstructured outdoor environments poses an even greater challenge.
Annotated training data is hard to obtain, as RGB\=/D cameras are not able to provide data at distances of more than \SI{10}{\meter} and their low resolution is not able to capture scenes crowded with differently sized objects.
Available datasets \citep{Geiger13,Cordts16,Huang18} use measurements from \gls{lidar} sensors to accumulate depth maps as ground-truth, which are, however, naturally sparse.
They can serve as an additional input for depth completion \citep{Cheng18} or as labels for actual depth prediction.
Utilizing stereo image pairs and a photo-consistency loss for semi-supervised training to estimate disparities is another option used in recent approaches \citep{Godard17,Yang18,Kuznietsov17}.
The application of \gls{side} in \glspl{adas} and \glspl{av} requires high precision and robustness.
Approaches to this challenging task have been proposed \citep{Gan18,Kendall18,Smolyanskiy18,Fu18,Guo18b,Li18b,Kong19,Li18c,Zhang18b}, but still require improvement for the application in \gls{ad} \citep{Smolyanskiy18}.

Estimating continuous depth values is a typical regression task.
However, by discretizing depth space into intervals, it can be cast as a classification problem \citep{Cao18,Gurram18,Fu18}.
While this is less intuitive, classification methods have been found to converge faster and more reliably.
This was shown by \citet{Fu18}, who advanced this approach by taking into account the ordinal characteristic of depth intervals and achieved top-ranking results in the \gls{kside} benchmark \citep{Uhrig17,Geiger13}.
Combining the properties of both tasks, \ie depth regression and classification of depth intervals, in order to exploit their individual advantages yields a \gls{mt} problem.

\Gls{mt} learning \citep{Caruana93,Caruana97} enables training of \glspl{cnn} that produce multiple outputs in a single round of inference.
In his review article, \citet{Ruder17} gives an extensive overview of \gls{cnn}-based \gls{mt} learning.
Driven by advances in methodology \citep{Sener18,Kendall18,Mallya18,Guo18,Zhao18}, \gls{mt} learning has become increasingly popular in computer vision.
It has successfully been applied to numerous applications.
In \gls{rsu}, problems that have been tackled using \gls{mt} learning include object detection and bounding box regression \citep{Teichmann18,Chen17,Chabot17}, as well as \gls{side} in combination with surface normal estimation or semantic segmentation \citep{Qi18,Zhang18,Eigen15,Ren18,Yang18,Kendall18}.

A different approach to employing \gls{mt} learning is the utilization of auxiliary tasks \citep{Liebel18,Chennupati19,Xu18} that merely serve as additional supervision to the network during training and are discarded during test-time.
This approach can be seen as an extension to comprehensive regularization terms in loss functions as used by \citet{Li18}.
It could be shown that by adding auxiliary tasks to a network the performance of the main task increases \citep{Liebel18,Chennupati19}.

Considering this prior work, we approach \gls{side} by posing it as a \gls{mt} problem with a main regression task and an auxiliary classification task.
As both tasks use depth measurements as ground-truth, with minor pre-processing applied in order to segment the continuous depth space into intervals for classification, the auxiliary supervisory signal does not require additional annotations.
By adding the auxiliary classification task as a regularizer, we expect training to converge faster and yield better results.
Closely related to the idea of casting \gls{side} to a classification task is the \gls{dorn}, proposed by \citet{Fu18}.
They do, however, use ordinal regression instead of classification and, furthermore, do not treat it as an additional task.
\citet{Kendall18} propose uncertainty-based weighting of individual tasks, which we build upon, but do not make use of auxiliary tasks.
Auxiliary tasks have been utilized before \citep{Liebel18,Chennupati19,Xu18}, however not with posing the same task in two different ways.
In contrast to \citet{Gurram18}, who use depth interval classification as pre-training, we train for regression and classification in an end-to-end manner.

The main contribution of this paper is the proposal of \gls{md}, a novel \gls{mt} approach to \gls{side} which incorporates both regression and classification.
This training strategy facilitates fast and robust convergence.
We, furthermore, provide an implementation of the proposed approach based on \gls{psp} \citep{Zhao17} and uncertainty-based weighting \citep{Kendall18} that we used to show the superiority of training with an auxiliary task as compared to basic regression.

\begin{figure*}
  \resizebox{\linewidth}{!}{%
      \includegraphics{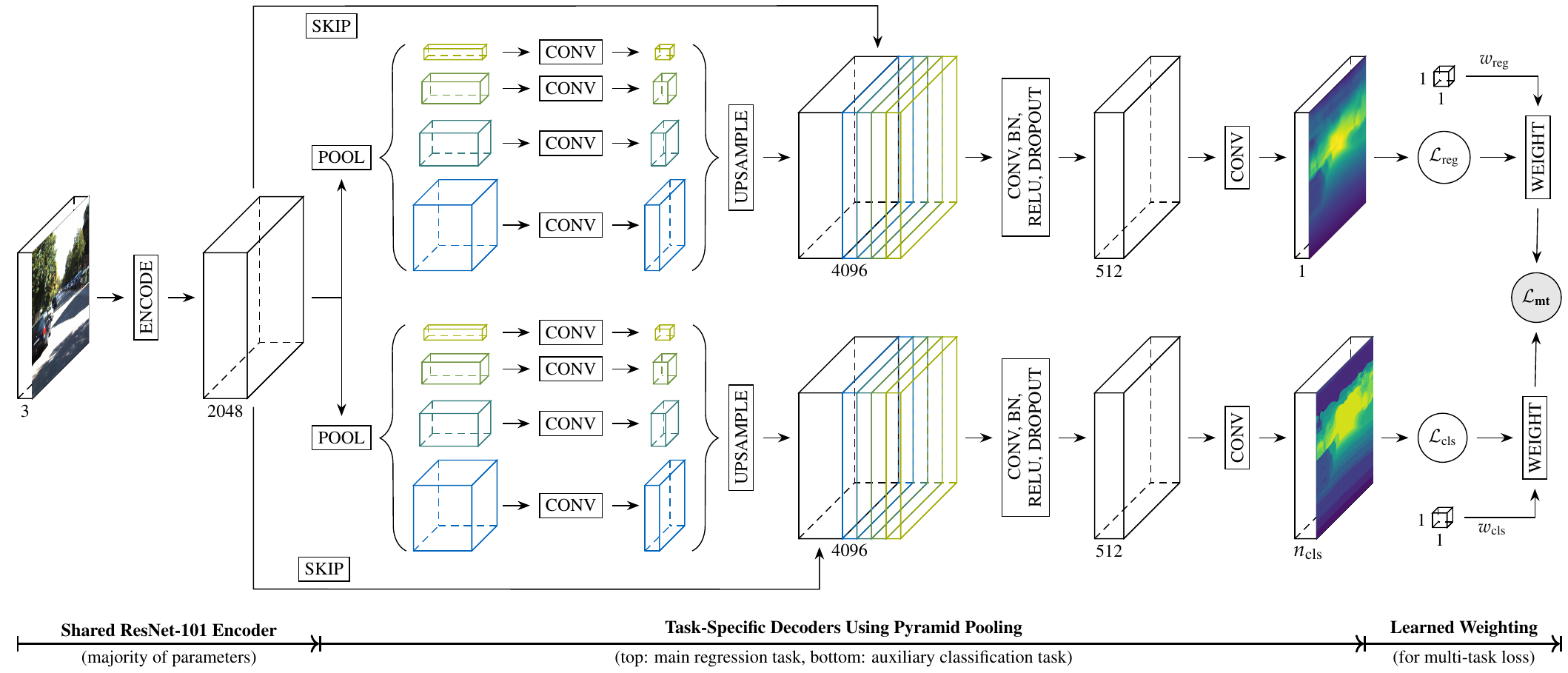}%
  }%
  \caption{Network architecture for the proposed MultiDepth approach featuring a shared ResNet-101 encoder with dilated convolutions and task-specific decoders including pyramid pooling for the main regression and the auxiliary classification task. The multi-task loss is constructed from the contributing single-task losses utilizing learned weighting terms, which represent task uncertainties.}
  \label{fig:network_architecture}
\end{figure*}


\section{Depth Prediction Using a Multi-Task Regression and Classification Loss}

Predicting depth maps from single images using a \gls{cnn} requires learning to estimate continuous depth values for each pixel.
This makes \gls{side} a prime example for regression tasks.
A common approach to this problem is, thus, optimizing a \gls{cnn} by the means of \gls{mse} or one of its variants.
However, it has been found that the convergence is slow and tends to yield suboptimal results \citep{Fu18}.

In contrast to this, \gls{side} can be posed as a classification problem using quantized depth values \citep{Cao18,Gurram18,Fu18}.
This allows employing state-of-the-art semantic segmentation methods.
Semantic segmentation, as another core task of many computer vision applications, such as \gls{ad}, has been successfully tackled in recent years.
State-of-the-art \gls{cnn} architectures for semantic segmentation \citep{Zhao17} most commonly utilize SoftMax cross entropy loss which is renowned for its stable convergence towards well-performing local minima.
Exploiting the advantageous properties of semantic segmentation methods for \gls{side}, however, comes at the price of inevitable quantization errors.

Combining the advantages of both, regression and classification, is a promising approach to the accurate estimation of depth values with stable convergence.
In order to achieve this goal, we propose \gls{md}, a multi-task training procedure for jointly learning two representations of a depth map, exemplarily shown in \cref{fig:intro_example}.
As network architectures for \gls{side} often use encoder-decoder structures, a second branch for the classification of depth intervals can easily be added as an auxiliary decoder \citep{Liebel18,Chennupati19}.
The auxiliary decoder is only active during training and can alternatively be disabled during test-time to make inference more time and memory efficient, or produce an additional redundant depth map that can be used to enhance or verify the regression result.
By sharing a major part of the network weights, a common and robust representation for both tasks is learned during training.
Especially the encoder part, which is expected to learn a rich representation of the input data, should be shared in order to encourage this behavior.
The decoders, on the other hand, only contain few layers to extract and reshape task-specific information from the shared encoded features.
State-of-the-art network architectures often use very deep classification networks as a backbone for deep feature extraction in the encoder, which, therefore, usually accounts for a vast majority of the parameters.

\subsection{Efficient Optimization of Depth in Log-Space}

In the context of \gls{ad}, even small errors in distance estimates for objects close to the camera---and hence the vehicle---may prove fatal.
Distances for faraway objects, on the other hand, only need to be estimated with lower accuracy.
This requirement can be considered by estimation in non-linear space.
Modern methods for \gls{cnn}-based \gls{side} use sophisticated loss functions in order to incorporate such desired properties.
One example of an objective function that considers a non-linear weighting of errors in log-space \citep{Eigen14} is commonly used as a metric for scoring depth estimation results.
We use the \gls{silog}
\begin{align}
  \text{SILog}(\M{d}, \M{d}^\star) = & \frac{1}{n} \sum_{i, j} (\log{d_{i, j}} - \log{d_{i, j}^\star})^2 \nonumber \\
                             & - \frac{1}{n^2} \left( \sum_{i, j} \log{d_{i, j}} - \log{d_{i, j}^\star} \right)^2
\end{align}
for comparing predicted depth maps $\M{d}^\star = ({d^\star_{i, j}})$ with $n$ depth values to the ground-truth $\M{d}$.
Utilizing extensive loss functions for training often yields superior results over more basic ones, such as the \gls{mse}.
The latter, however, are less computationally expensive thus speeding up the training process.
While they might not directly optimize the final objective, they act as a proxy of reasonable quality \citep{Goodfellow16}.

In order to overcome the limitations of both approaches, we relocate computationally expensive operations to the data pre-processing stage.
Thanks to parallel pre-loading of batches, this allows us to optimize our objective more directly while still maintaining the speed and efficiency of traditional loss functions.
Hence, we transform depth values $d$ in our training data to normalized log-space during data loading.
After globally defining the upper and lower boundaries $d_\text{min}$ and $d_\text{max}$, we calculate the transformed depth as
\begin{align}
  d_\text{log} = \frac{
                  \log_\e \left( d - d_\text{min} + 1 \right)
                }{
                  \log_\e \left( d_\text{max} - d_\text{min} + 1 \right)
                }
                \quad \text{.}
  \label{eq:encode_depth}
\end{align}

\subsection{Regression of Depth Values}

In order to extract the desired information from the encoder features, task-specific decoders can be added to the network.
For the main task, \ie regression of continuous depth values, the decoder directly predicts values in normalized log-space.
Few network layers to condense and reshape information along all dimensions of the feature map are expected to suffice here, given the rich representation of the input data provided by the encoder.

\subsection{Classification of Depth Intervals}

The auxiliary task, \ie classification of discrete depth ranges, requires corresponding target labels.
In order to partition the continuous depth space, we set a number of intervals $n_\text{cls}$.
Choosing a relatively low number of classes ensures stable convergence while still preserving enough detail to support the regression part of the network with basic depth perception.
While \citet{Fu18} report 80 intervals to be the optimum for their ordinal regression method, this is not directly applicable to our approach in which the classification task merely supports the regression branch of the network.
Hence, trading high quantization errors for stable convergence and robust performance is desirable in our case.
Following our general concept of estimating non-linearly scaled depth values and existing approaches \citep{Fu18}, we set our intervals to be uniformly distributed in normalized log-space.

Apart from different target labels and a necessary change in the output feature dimension from one to $n_\text{cls}$, the decoder for the classification task closely resembles the structure of the regression branch.
Keeping the task-specific branches shallow, and thus the ratio of shared \vs individual parameters high encourages the learning of high-level features that are suitable for both tasks.

\subsection{Multi-Task Optimization}
\label{subsec:multi_task}

Jointly training both tasks requires combining the individual losses $\loss_\text{task}$ to a single multi-task objective $\loss_\text{mt}$, subject to optimization.
A na\"ive approach to this is simply summing up $\loss_\text{mt} = \loss_\text{reg} + \loss_\text{cls}$.
The range and variance of values produced by $\loss_\text{reg}$ and $\loss_\text{cls}$ may differ significantly, especially since they belong to different families of loss functions.
Hence, weighting with $w_\text{reg}$ and $w_\text{cls}$, such that $\loss_\text{mt} = w_\text{reg} \cdot \loss_\text{reg} + w_\text{cls} \cdot \loss_\text{cls}$ allows for adjusting the contribution of each loss.
One possible way of finding appropriate values for the weights is treating them as hyperparameters to be manually tuned.
This is a tedious process that results in a set of constant weights that might be suitable in general, but fail to adapt to changes during training.
Therefore, automatically optimizing a set of dynamic parameters is a promising approach.

Existing methods propose to weight tasks with respect to their homoscedastic uncertainty \citep{Kendall18} or difficulty \citep{Guo18}.
We follow the former and introduce weighting parameters $w_\text{reg} = 0.5 \cdot \exp(-s_\text{reg}^2)$ and $w_\text{cls} = \exp(-s_\text{cls}^2)$ with $s_\text{task} = \log_\e(\sigma_\text{task}^2)$.
While $\sigma_\text{task}$ represent the actual task uncertainties, we optimize for $s_\text{task}$, due to numerical stability, as advised by the original authors \citep{Kendall18}.
Since minimizing $\loss_\text{mt} = w_\text{reg} \cdot \loss_\text{reg} + w_\text{cls} \cdot \loss_\text{cls}$ favors the trivial solution $w_\text{reg} = w_\text{cls} = 0$, they furthermore propose to add regularization terms to the weighted single-task losses.
Adding such terms $r_\text{task} = 0.5 \cdot s_\text{task}$ yields our final multi-task loss
\begin{align}
  \loss_\text{mt} = w_\text{reg} \cdot \loss_\text{reg} + r_\text{reg} + w_\text{cls} \cdot \loss_\text{cls} + r_\text{cls} \text{\quad .}
  \label{eq:mt_loss}
\end{align}

\section{Experiments}

To evaluate our approach, we implemented a deep \gls{cnn} and trained it on a large \gls{rsu} dataset.

\subsection{Network Implementation}

As networks for \gls{side} and semantic segmentation share typical properties, such as their feed-forward auto-encoder structure, architectures are often shared across both tasks \citep{Laina16,Li17}.
Furthermore, our auxiliary pixel-wise depth interval classification task, in fact, is a segmentation task.
Therefore, we deem well-performing deep \gls{cnn} architectures for semantic segmentation suitable as a basis for our implementation.

With its proposed \gls{pp} module, the \gls{psp} \citep{Zhao17} is able to accumulate information from spatial context.
The very deep ResNet-101 \citep{He16} used as a backbone for the feature encoder, on the other hand, provides the necessary capacity to learn a rich and robust representation of the input data.
The network also features a natural encoder-decoder structure using \gls{pp} and dilated convolutions \citep{Chen14,Yu16} to preserve high-resolution feature maps up to the output.
As a result, the \gls{psp} achieves state-of-the-art results on various challenging semantic segmentation datasets \citep{Zhao17}.

Since these properties match with our requirements, we re-implemented their network architecture in PyTorch and adapted it to our \gls{mt} approach.
An overview of our final \gls{md} network architecture is shown in \cref{fig:network_architecture}.
The auxiliary loss used in the original implementation of \citet{Zhao17} was not implemented.

Parameters between both tasks are shared through the ResNet encoder, which accounts for approximately 50\% of the total number of parameters in the network.
Sharing a majority of parameters between both tasks enforces the learning of a common representation in the 2048-dimensional feature map.
Since both tasks represent a perception of depth, we expect these features to already contain high-level information that strongly hints towards the final outputs.
This representation needs to be condensed for each task.
The weights of the \gls{pp} modules, the following blocks containing a single convolution (along with batch normalization, ReLU activation, and dropout), and the final convolutional layers for reshaping the intermediate representation to the required output format are task-specific.
We use skip connections as in the original \gls{psp} to retain spatial structure.

\begin{figure}
  \centering
  \scriptsize

  \begin{subfigure}[t]{.48\linewidth}
    \includegraphics[width=\linewidth]{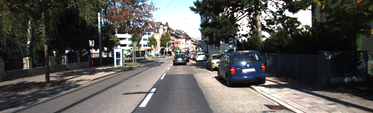}
    \vspace{-1mm}
    \caption{}
    \label{subfig:example_kitti_rgb}
  \end{subfigure}
  \hfill
  \begin{subfigure}[t]{.48\linewidth}
    \scriptsize
    \fboxsep = 0mm  
    \fboxrule= .5pt 
    \fbox{\includegraphics[width=\linewidth - \fboxrule * 2]{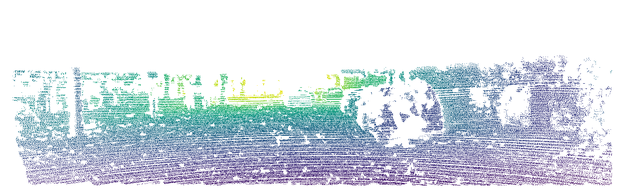}} \\
    \scriptsize
    \SI{5}{\meter} \hfill
    \raisebox{0.35mm}{\includegraphics[width=.7\linewidth, height=0.75mm]{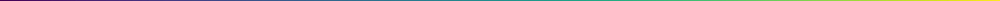}}
    \hfill \SI{80}{\meter}
    \vspace{-0.1cm} 
    \caption{}
    \label{fig:example_kitti_gt}
  \end{subfigure}

  \vspace{0.15cm}

  \begin{subfigure}{\linewidth}
    \includegraphics[width=\linewidth]{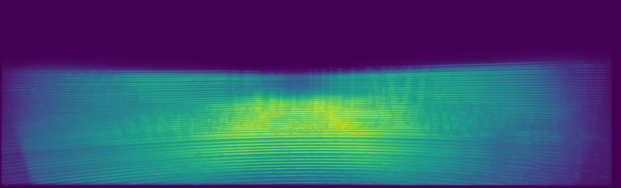} \\
    \scriptsize
    Ground-truth available: \hfill 0\%
    \raisebox{0.35mm}{\includegraphics[width=.575\linewidth, height=0.75mm]{figures/example_kitti_colorbar}}
    52\%
    \vspace{-0.1cm} 
    \caption{}
    \label{subfig:depth_distribution_kitti_spatial}
  \end{subfigure}

  \vspace{2mm} 

  \begin{subfigure}{\linewidth}
    \includegraphics{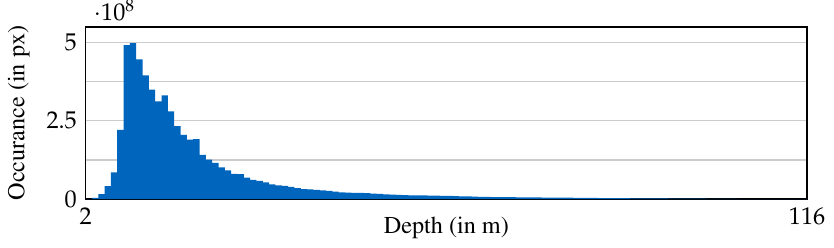}
    \vspace{-0.2cm} 
    \caption{}
    \label{subfig:depth_distribution_kitti_depth}
  \end{subfigure}

  \caption{Sample from the KITTI depth prediction dataset \citep{Uhrig17, Geiger13} with RGB image (a) and sparse ground-truth depth (b).
  Distribution of (c) available ground-truth depth values per pixel in the training set with an apparent lack of measurements in the upper part of the images, and (d) of valid depth values in the training set.}
  \vspace{-0.5cm}
\end{figure}

\subsection{Training and Validation Data}
\label{subsec:data}

Predicting depth with high robustness, redundancy, and accuracy is especially important for \gls{ad}.
As \gls{md} is designed to fulfill these conditions, we chose to train and evaluate our method on images of road scenes.
Amongst the most popular and comprehensive datasets for \gls{rsu} is the \gls{kitti} dataset \citep{Geiger13}, which provides ground-truth for several highly relevant tasks.
The \gls{kside} dataset \citep{Uhrig17, Geiger13} provides sparse point clouds, acquired using \gls{lidar} sensors.
The dataset contains roughly \num{86000} training samples.
Ground-truth depth values, as exemplarily shown in \cref{fig:example_kitti_gt}, are available for roughly 12\% of the pixels in each image on average.
For some samples, however, this number drops to only 0.8\%.
\Cref{subfig:depth_distribution_kitti_spatial} shows the unequal spatial distribution of depth measurements with a significant lack of data for the upper part of the images.

We train on the full set of training images and validate on the \num{1000} images of the provided ``val\_selection\_cropped'' set.
Even though the evaluation of our approach is, naturally, largely based on relative comparisons and ablation studies, we still want to evaluate \gls{md} on a challenging dataset with respect to the state of the art.
Hence, we score our final results on the official \gls{kside} benchmark \citep{Uhrig17}.

\subsection{Data Augmentation and Scaling}

To produce appropriately sized input images for our network, we implemented random cropping with a resolution of \SI{128x128}{\pixel} to \SI{256x256}{\pixel}, depending on the respective experiment.
While both of the used patch sizes are relatively large, using smaller patches leads to an increasing number of images with a distinct lack of ground-truth information (\cf \cref{subfig:depth_distribution_kitti_spatial}).
We refrained from scaling images as a data augmentation measure.
This allows us to exploit the fully-convolutional nature of our network during inference since the size of objects in the cropped patches does not differ from the original full-sized images.
Counterintuitively, horizontal flipping appears to improve results even for road scenes, where differences in the left and right side of the images are systematic and meaningful \citep{Romera18}.

Depth values were scaled according to \cref{eq:encode_depth} with $d_\text{min} = \SI{2}{\meter}$ and $d_\text{max} = \SI{125}{\meter}$, following the distribution of depth values in the training dataset (\cf \cref{subfig:depth_distribution_kitti_depth}).
While the lower bound was tightly set, in order not to waste accuracy in the critical lower part of the normalized log-space, the upper bound is more generous, to allow for higher depth values which account for very little accuracy in the normalized log-space.
Class-labels for the classification branch of the network were derived by binning the scaled depth values to intervals in $\left[ d_\text{cmin}, d_\text{cmax}\right]$, equally spaced in normalized log-space.
By setting the lower and upper clipping planes $d_\text{cmin}$ and $d_\text{cmax}$ for the quantization to $d_\text{cmin} > d_\text{min}$ and $d_\text{cmax} < d_\text{max}$, we simplify the auxiliary task.
As the classification task is only meant to support training, this helps to keep the auxiliary task sufficiently easy to train.

\subsection{Training Procedure}

\begin{figure}
  \includegraphics{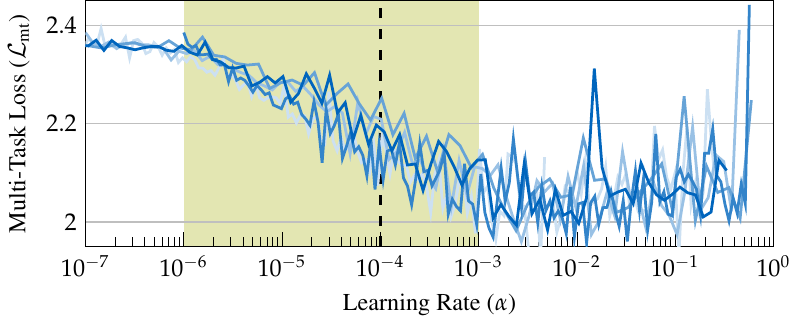}
  \caption{Experimental estimation of a suitable learning rate. The shaded interval of learning rates decreases the loss during training, the dashed line marks the selected learning rate.}
  \label{fig:hyperparameter_tuning_lr}
\end{figure}

\begin{figure}
  \includegraphics{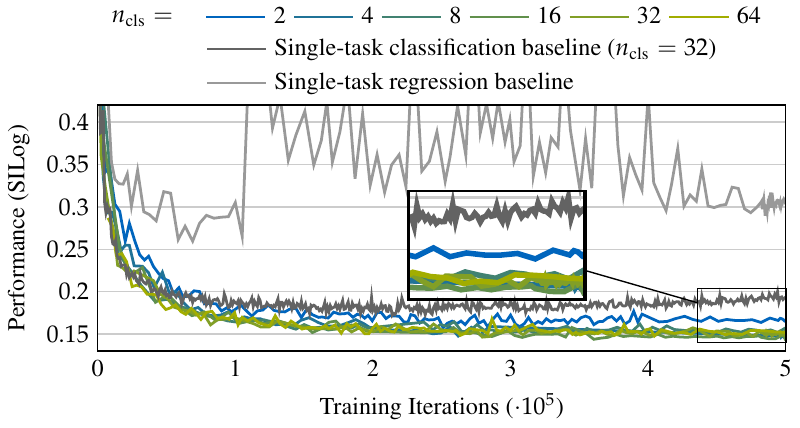}
  \caption{Validation results for single-task regression and classification \vs multi-task training with different $n_\text{cls}$ for the auxiliary classification task.
  The regression baseline diverges quickly at first and converges to an undesirable local minimum afterward.
  As expected, the classification baseline converges fast and stable.
  Multi-task training with $n_\text{cls} \geq 2$ considerably improves performance over both baselines.
  Best results achieved for $n_\text{cls} \geq 4$.}
  \label{fig:ablation_nclasses}
\end{figure}

\begin{figure*}
  \includegraphics{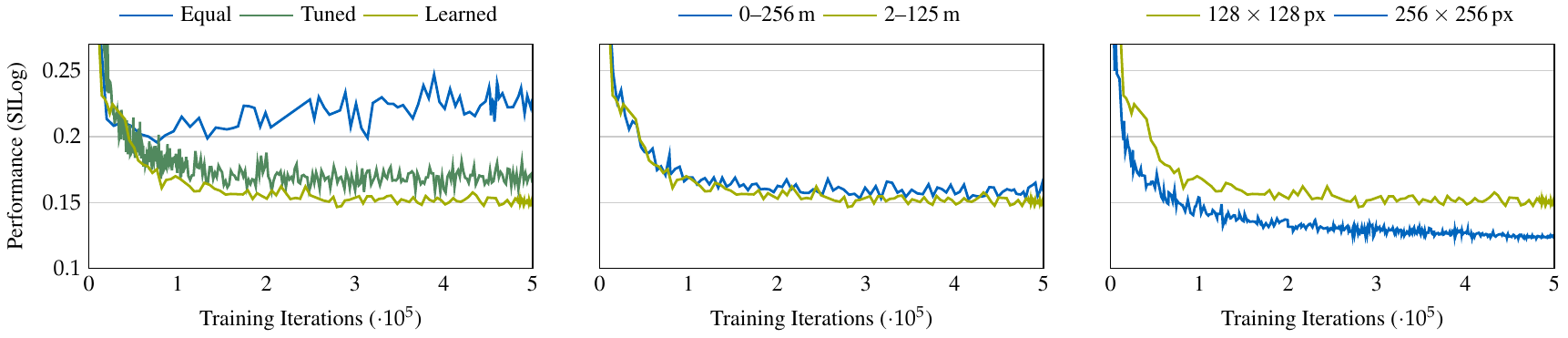}
  \vspace{-.5cm}

  \hspace{1cm}
  \begin{subfigure}[t]{4.95cm}
    \caption{}
    \label{subfig:ablation_mt}
  \end{subfigure}
  \hfill
  \begin{subfigure}[t]{4.95cm}
    \caption{}
    \label{subfig:ablation_scaling}
  \end{subfigure}
  \hfill
  \begin{subfigure}[t]{4.95cm}
    \caption{}
    \label{subfig:ablation_patchsize}
  \end{subfigure}
  \hspace{0.25mm}

  \vspace{-0.25cm} 
  \caption{Validation results using different settings for (a) multi-task weights, (b) depth value scaling, and (c) patch size. Models using learned weighting, normalization bounds adapted to the data distribution, and large patches for training yield best results with weighting being the most important factor.}
  \label{fig:ablation_studies}

\end{figure*}

\begin{figure}[t!]
 \centering
 \includegraphics{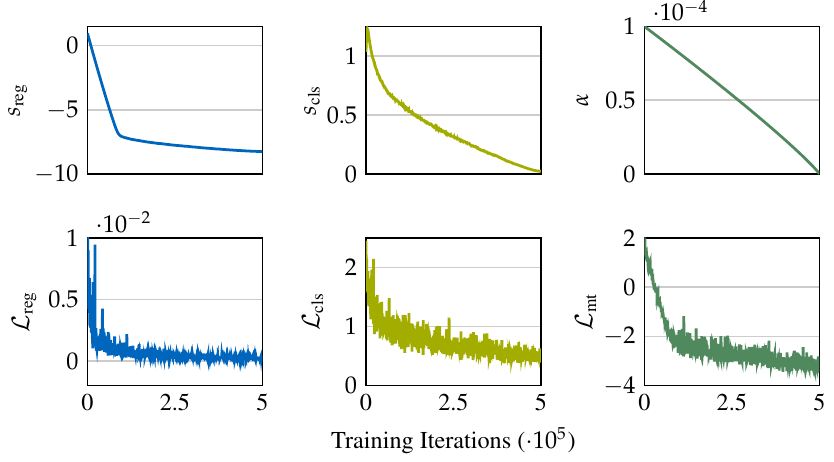} \\
 \includegraphics{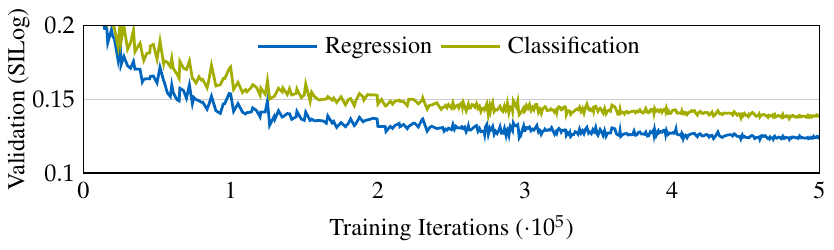}

 \caption{Training results for the final model configuration showing an initial phase of adjusting the weights $s_\text{task}$ according to the uncertainty of $\loss_\text{task}$ followed by a continuous decrease of both. The combined $\loss_\text{mt}$ is optimized using a decaying $\alpha$. Validation results show the stable convergence of both outputs with the regression yielding superior results due to quantization errors in the auxiliary classification output.}
 \label{fig:training_results}
\end{figure}

Since ground-truth for both of our tasks is sparse, we implemented a sparse \gls{mse} loss for the regression task and a sparse SoftMax cross entropy loss for the classification task.
The multi-task loss, which is subject to optimization, was constructed as presented in \cref{eq:mt_loss}.
The weighting parameters for the multi-task loss were initialized as $s_\text{reg} = s_\text{cls} = 1$ and optimized with the network parameters.
\Citet{Kendall18} point out the robustness of the weighting parameters with regard to their initialization, which our experiments confirmed.

Settings for batch and patch sizes are coupled and bound by each other in terms of GPU memory consumption.
Due to the sparsity of the ground-truth data, larger patches are favorable here.
The optimization procedure, on the other hand, generally benefits from larger batch sizes.
Using a batch size of \num{32} for our experiments with a patch size of \SI{128x128}{\pixel} realized the best trade-off in our experiments.
We used Adam \citep{Kingma15} with $\beta_1 = 0.9$ and $\beta_2 = 0.999$ and a weight decay $\lambda = 0.0001$ for optimization.

The probably most crucial hyperparameter for the optimization of a deep \gls{cnn} is the \gls{lr} $\alpha$ \citep{Smith17}.
We followed the approach of \citet{Smith17}, who propose to run training for few iterations, \ie optimization steps, with increasing learning rate.
By starting from minuscule learning rates and exponentially increasing them to excessively high values, we cover the full space of realistic $\alpha$ for training with only \num{250} steps of optimization.
Repeating this experiment starting from different $\alpha$ and using varying parameters for the exponential scheduler, we obtained an $\alpha$ \vs $\loss_\text{mt}$ diagram, shown in \cref{fig:hyperparameter_tuning_lr}.
Four typical intervals of $\alpha$ can be observed.
Too conservative settings with $\alpha < 10^{-6}$ result in a stagnation of $\loss_\text{mt}$.
The second interval of values $10^{-6} < \alpha < 10^{-3}$ yields a decrease of $\loss_\text{mt}$, marking the desired range.
When surpassing this interval, $\loss_\text{mt}$ no longer decreases and finally diverges for $\alpha > 10^{-1}$.
From these findings, we deem learning rates in the second interval as suitable and selected $\alpha = 10^{-4}$, which is close to the steepest decent of $\loss_\text{mt}$, for our experiments.
We applied a polynomial scheduler with $\gamma = 0.9$ to slowly decay $\alpha$ to zero over $5 \cdot 10^5$ training iterations, which roughly equals \num{90} epochs.
Note that due to random cropping of patches from the input data, the number of epochs is not a very well-suited measure here.

\subsection{Ablation Studies}
\label{subsec:ablation_studies}

Evaluation of our approach with learned and dynamic weighting between the differently posed depth estimation tasks that mutually support each other requires a careful protocol.
To analyze the proposed approach, we conducted a number of ablation studies and present the results in the following subsections.
Training was conducted from scratch without using pre-trained weights for the ResNet encoder.
Networks were trained from scratch on a simple desktop machine, equipped with a single NVIDIA GeForce 1080Ti GPU, in approximately 5 days.

\subsubsection{Number of Classes}
\label{subsec:ablation_nclasses}

To obtain a baseline result, we trained the regression branch of our network without the additional classification branch.
As seen in \cref{fig:ablation_nclasses}, the single-task results get better at first but diverge after iteration $10^5$.
Convergence towards an undesirable local minimum using vanilla regression has also been reported by \citet{Fu18}.
We also trained a model for single-task classification by removing the regression branch from the network architecture.
As expected, casting \gls{side} to a depth interval classification task eliminates the problem of slow and unstable convergence and yields much better results overall.

Following our \gls{md} approach by adding an auxiliary classification branch to the regression network helps to stabilize training, even when using as little as two classes.
This result is related to the observations of \citet{Li18}, who include a term for foreground/background separation in their loss function.
Increasing the number of classes further improves stability during training and the overall performance.
However, using $n_\text{cls} \geq 4$ only yields minor improvements.
Another notable outcome is that using more classes than necessary does not significantly degrade results.
It does, however, yield a secondary output that suffers from far lower quantization errors and could, thus, potentially be used as a redundant signal.
An example of both outputs is shown in \cref{fig:intro_example}.
Models with $4 \leq n_\text{cls} \leq 64$ achieved almost identical results on the validation set.
We selected $n_\text{cls} = 32$ for further experiments due to best convergence.

\begin{figure*}
  \includegraphics[width=.24\linewidth]{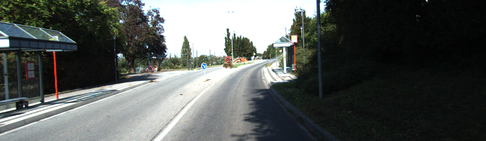} \hfill
  \includegraphics[width=.24\linewidth]{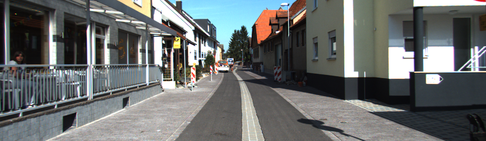} \hfill
  \includegraphics[width=.24\linewidth]{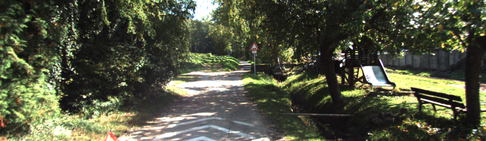} \hfill
  \includegraphics[width=.24\linewidth]{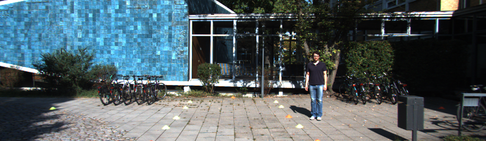} \\[1mm]
  \includegraphics[width=.24\linewidth]{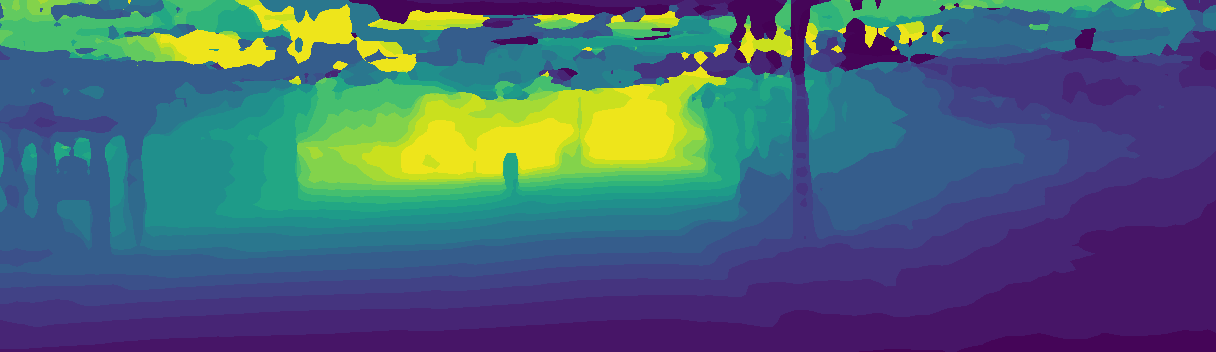} \hfill
  \includegraphics[width=.24\linewidth]{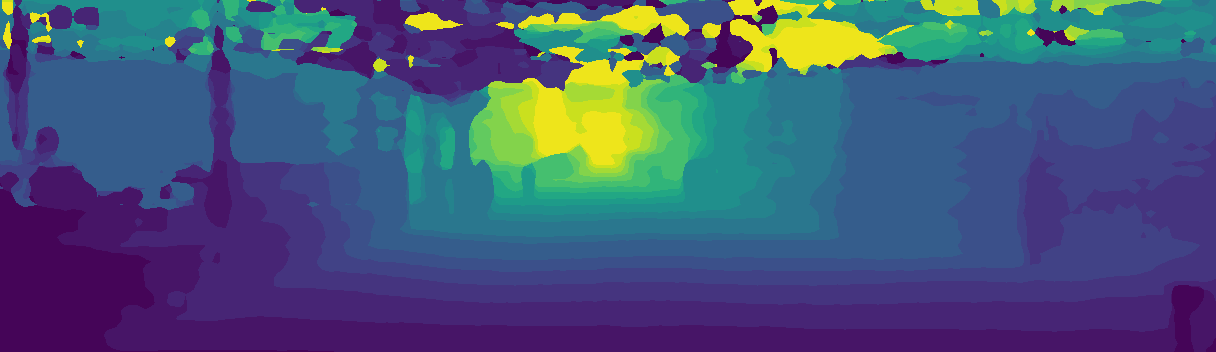} \hfill
  \includegraphics[width=.24\linewidth]{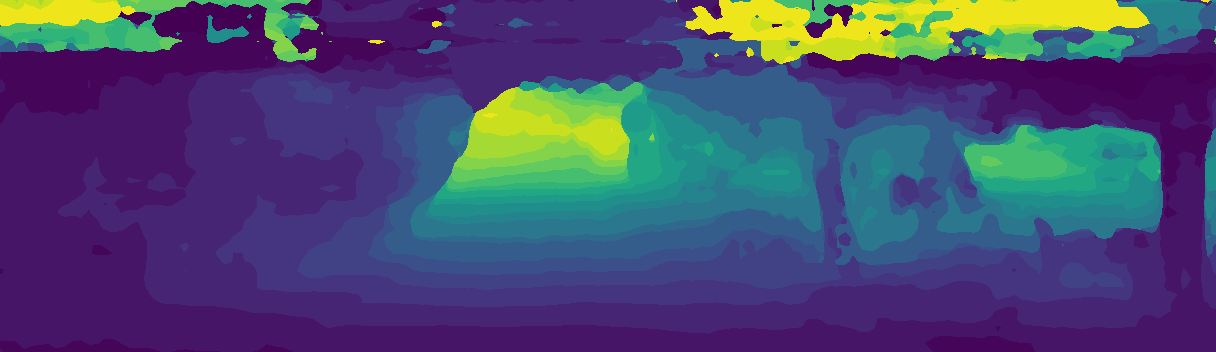} \hfill
  \includegraphics[width=.24\linewidth]{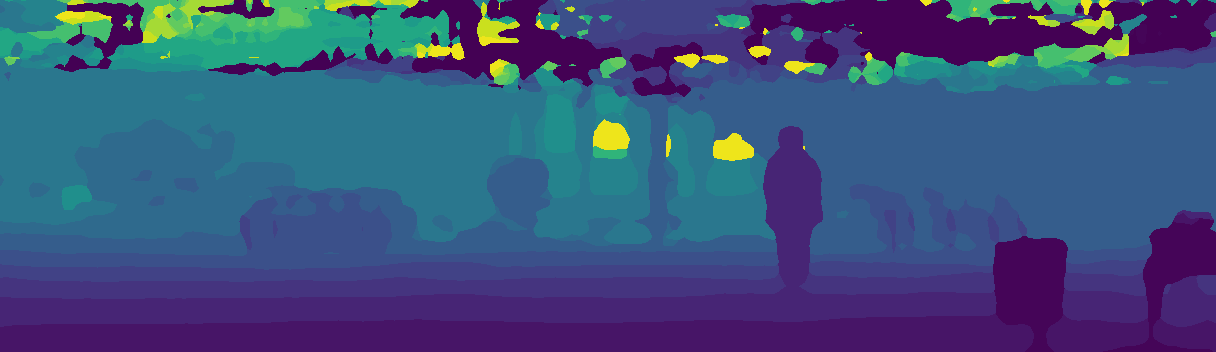} \\[1mm]
  \includegraphics[width=.24\linewidth]{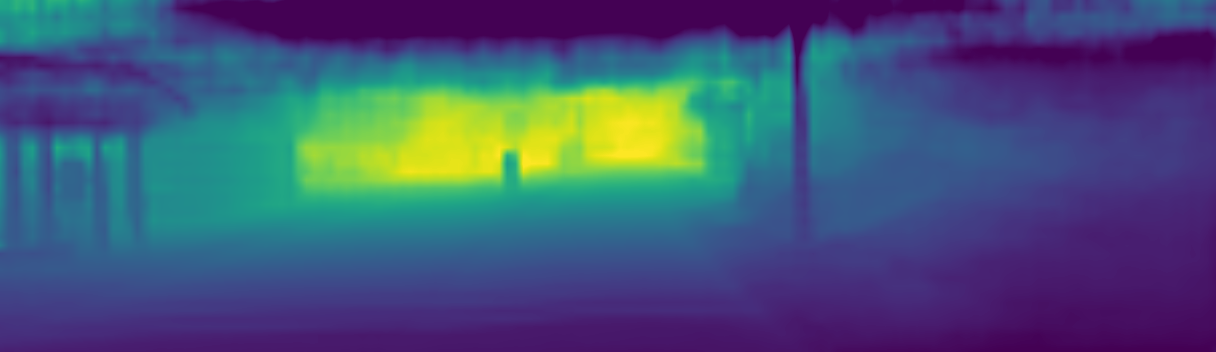} \hfill
  \includegraphics[width=.24\linewidth]{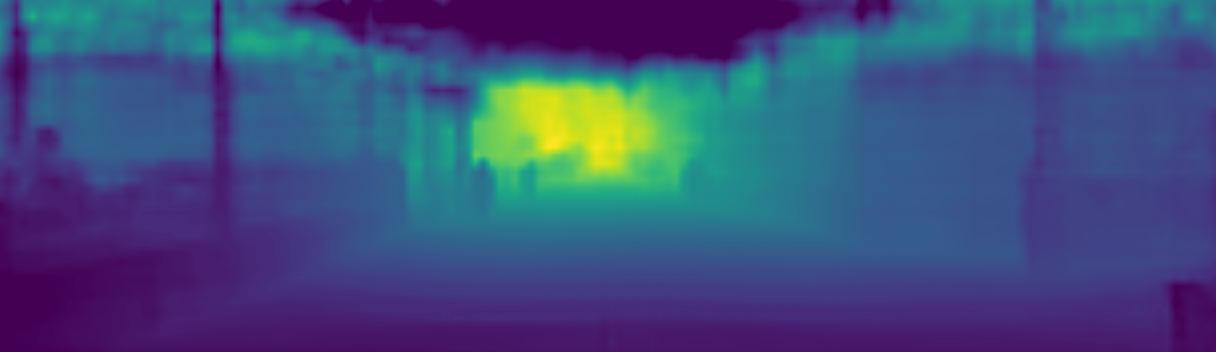} \hfill
  \includegraphics[width=.24\linewidth]{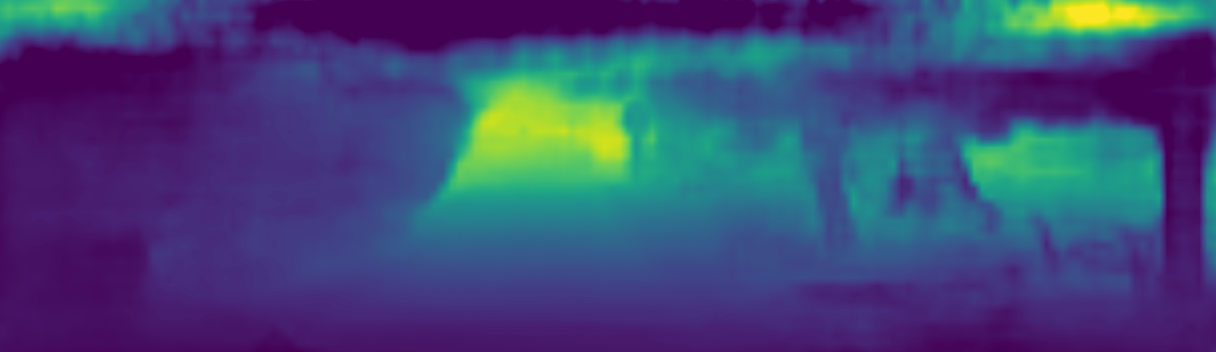} \hfill
  \includegraphics[width=.24\linewidth]{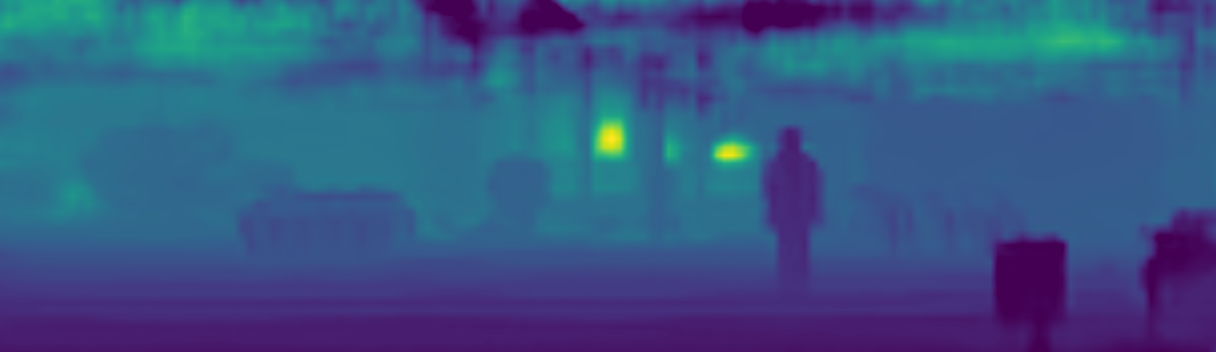} \\[1mm]
  \includegraphics[width=.24\linewidth]{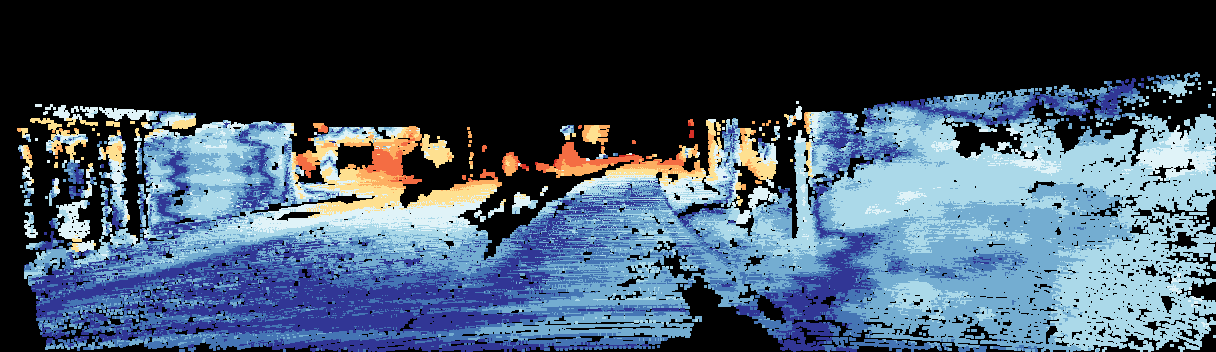} \hfill
  \includegraphics[width=.24\linewidth]{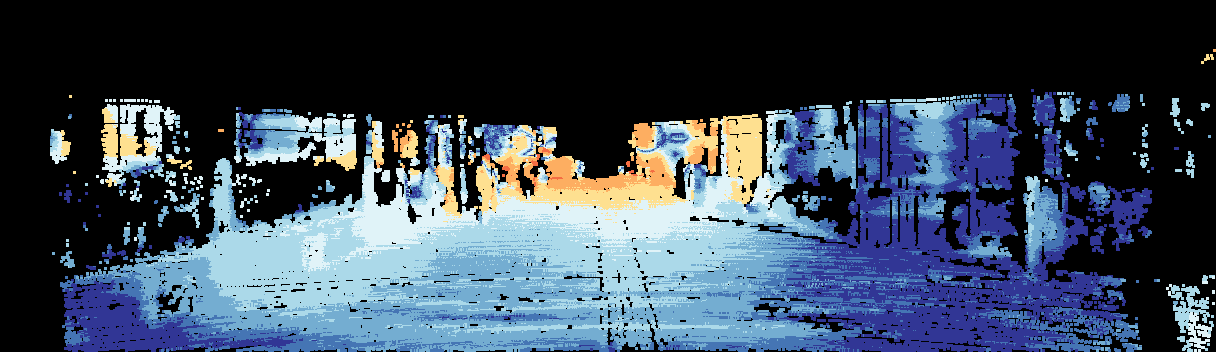} \hfill
  \includegraphics[width=.24\linewidth]{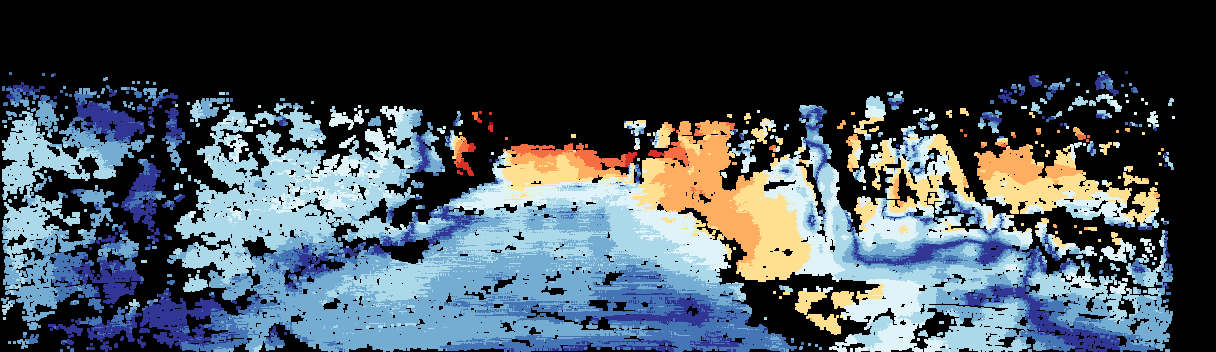} \hfill
  \includegraphics[width=.24\linewidth]{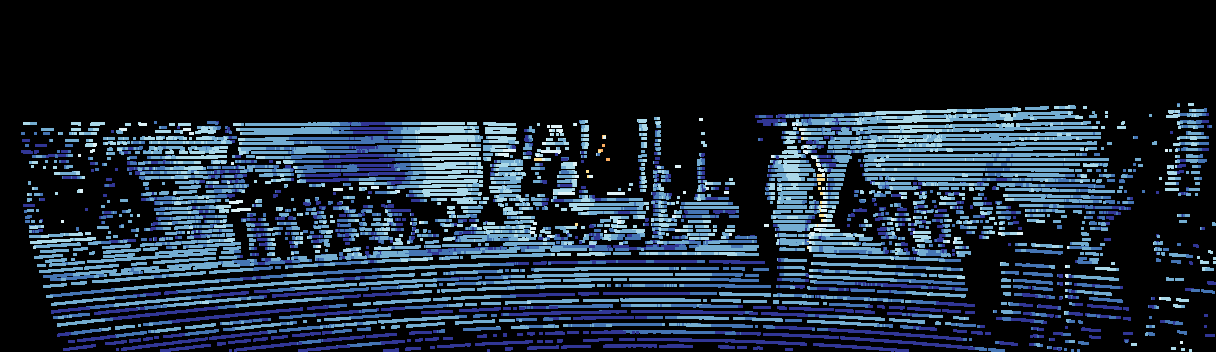}
  \caption{Qualitative evaluation of our estimation results for unseen KITTI depth prediction test images \citep{Uhrig17, Geiger13} (first row) with results of the auxiliary classification task (second row), results of the main regression task (third row) and color coded error images of the regression result compared to the sparse \gls{lidar} point cloud (fourth row).}
  \label{fig:result_images}
\end{figure*}

\subsubsection{Multi-Task Weighting}
\label{subsec:ablation_mt}

In order to evaluate the influence of learned weights \citep{Kendall18} on the training process, we ran additional experiments with equally weighted tasks and manually tuned weighting terms.
\Cref{subfig:ablation_mt} shows the training progress, evaluated on the validation set, for configurations with $n_\text{cls} = 32$ (\cf \cref{subsec:ablation_nclasses}).
The baseline with fixed and equal weights for both tasks converges to an unstable local minimum and its best result is significantly worse than the performance of the model trained with learned weights.
Manually tuned pairs of fixed weights led to much better results compared to the equally weighted baseline, but still perform significantly worse than the learned dynamic weighting, as seen in \cref{tab:results_multidepth}.
In our experiments, learned sets of weights always outperformed manually tuned weights, showing the effectiveness of this technique.
Note that manual tuning becomes increasingly complex when adding more tasks \citep{Kendall18,Liebel18}, but even for the limited set of tasks used in our experiments, learning the weighting terms saves precious computation time while introducing as little as two additional parameters to the optimization problem.

\subsubsection{Log-Normalization of Depth Values}
\label{subsec:ablation_scaling}

In the data pre-processing stage, which was done online in parallel threads, we scaled the depth values according to \cref{eq:encode_depth}.
We conducted dedicated experiments on how setting the limits $d_\text{min}$ and $d_\text{max}$ for the normalization influences the quality of the final results.
We set them according to the minimum and maximum depth present in the dataset (\cf \cref{subfig:depth_distribution_kitti_depth}), which is a good estimate of real-world conditions, and compared them to scaling from \SIrange{0}{256}{\meter}, which are the theoretical bounds for the depth encoding in the \gls{kitti} data format.
Again, both training runs were conducted using $n_\text{cls} = 32$ (\cf \cref{subsec:ablation_nclasses}) and learned task weights (\cf \cref{subsec:ablation_mt}).
The former outperformed the latter, as scaling to the relevant range spreads the numerical precision with respect to data and application.
Note that this relevant range of depth values might differ for other applications, such as indoor settings.

\subsubsection{Patch Size}
\label{subsec:ablation_patchsize}

In a final study, we trained on different patch sizes.
As already noted earlier, this is especially relevant due to the sparse ground-truth data.
In order to show this, we trained on bigger patches of \SI{256x256}{\pixel} and report the results in \cref{subfig:ablation_patchsize}.
Since training with similar settings as in the preceding experiments but a bigger patch size is more memory intensive, we ran this experiment on two NVIDIA 1080Ti GPUs using parallelization with batch splitting and a batch size of \num{16}.
Training using bigger patches outperformed training on smaller ones, which was expected.
The results achieved using this configuration yielded the best results in our series of experiments and were thus submitted to the \gls{kside} benchmark for final scoring.

\begin{table}[t!]
  \vspace{-.5cm}
  \caption{Performance of models covering multiple aspects of MultiDepth on the validation set.
  Classification results given for $n_\text{cls} = 32$.
  Regression and classification predictions were inferred in a single pass for the MultiDepth models.
  All configurations (with a negligible exception) outperform the single-task baselines by far.}
  \label{tab:results_multidepth}
  \vspace{.25cm}
  \scriptsize
  \begin{tabularx}{\linewidth}{@{}lrl*{2}{S[table-format=2.2, table-number-alignment=right]}@{}}
    \toprule
    \multicolumn{3}{@{}l}{Experiment} & \multicolumn{2}{l}{Results (SILog)} \\
    \cmidrule(r){1-3} \cmidrule(l){4-5}
    \multicolumn{1}{@{}l}{Method} & \multicolumn{1}{l}{Patch Size} & \multicolumn{1}{l}{Weighting} & \multicolumn{1}{X}{Classification} & \multicolumn{1}{X}{Regression} \\
    \cmidrule(r){1-1} \cmidrule(lr){2-2} \cmidrule(lr){3-3} \cmidrule(lr){4-4} \cmidrule(l){5-5}
    Baseline   & \SI{128x128}{\pixel} & single-task        & \multicolumn{1}{r}{\textemdash} & 25.96                           \\
    Baseline   & \SI{128x128}{\pixel} & single-task        & 17.22                           & \multicolumn{1}{r}{\textemdash} \\
    MultiDepth & \SI{128x128}{\pixel} & equal              & 17.63                           & 19.56                           \\
    MultiDepth & \SI{128x128}{\pixel} & man. tuned         & 15.99                           & 15.75                           \\
    MultiDepth & \SI{128x128}{\pixel} & learned            & 15.62                           & 14.65                           \\
    MultiDepth & \SI{256x256}{\pixel} & learned            & 13.70                           & 12.27                           \\
    \bottomrule
  \end{tabularx}
  \vspace{-.5cm}
\end{table}

\subsection{Final Training Results}

Based on our experiments (see \cref{subsec:ablation_studies}), we selected the model with the highest validation score for submission to the \gls{kside} benchmark.
\Cref{tab:results_multidepth} lists the performance of the most relevant configurations.
\Cref{fig:training_results} shows the training progress including the task weights, the learning rate, single-task and multi-task losses, and results on the validation set for the outputs of the regression and classification branch.
Since the task uncertainties differ widely, as seen in $\loss_\text{reg}$ and $\loss_\text{cls}$, $s_\text{reg}$ quickly shifted to a suitable range (note that due to log-scaling in the weighting function $s_\text{reg} < 0$ corresponds to weights close to zero).
After a minor initial increase, $s_\text{cls}$ stabilizes and slowly decreases together with $s_\text{reg}$ over the course of training to account for the converging single-task losses.
Validation results show the expected superiority of the regression over the classification results due to quantization errors, good convergence, and no signs of overfitting.

Depth maps for the images of the anonymous test set were predicted using the regression branch of the final model and submitted to the \gls{kside} benchmark where they scored an \gls{silog} of \num{16.05}.
Top ranking methods achieve comparable to better scores (\gls{dorn} \citep{Fu18}: \num{11.77}, VGG16-UNet \citep{Guo18b}: \num{13.41}, HGR framework \citep{Zhang18b}: \num{15.47}).
Qualitative results, given in \cref{fig:result_images}, show that our network is able to infer useful depth maps overall.
Note that we did not implement any pre-training or post-processing in order to enhance the prediction results since this might interfere with convergence during training---thus rendering the observation of relative improvements made using our method impossible.

\section{Conclusion}

We presented \gls{md}, a method for introducing an auxiliary depth interval classification task to \gls{side} networks.
An implementation of our network has been applied to the challenging problem of \gls{side} in \gls{rsu}.
In extensive experiments, we showed the benefits of posing \gls{side} as a \gls{mt} problem with additional supervision through automatically derived ground-truth labels.

The single-task regression and classification baselines were outperformed by far using our method with 32 classes for the auxiliary classification branch and learned weighting between the tasks.
\gls{md} achieved an \gls{silog} of \num{16.05} on the anonymous \gls{kside} test set \citep{Uhrig17,Geiger13}, showing that our training strategy also yields good results overall.

Source code for all experiments presented in this paper is publicly available online%
\footnote{\url{https://github.com/lukasliebel/MultiDepth}}.


\AtNextBibliography{\small}
\printbibliography

\end{document}